\documentclass[11pt,a4paper]{article}

\usepackage[english]{babel}
\usepackage[utf8x]{inputenc}
\usepackage[T1]{fontenc}
\usepackage{graphicx}

\usepackage{naaclhlt2018}
\usepackage{times}
\usepackage{latexsym}

\usepackage{url}
\aclfinalcopy 

\title{An Exploration of State-of-the-art Methods for Offensive Language Detection}

\author{
  Harrison Uglow \\
  Imperial College of London \\
  {\tt hu115@ic.ac.uk} \\\And
    Martin Zlocha \\
  Imperial College of London \\
  {\tt mz4315@ic.ac.uk} \\\And
  Szymon Zmyslony \\
  Imperial College of London \\
  {\tt sz6315@ic.ac.uk} 
  }

\date{}

\begin{document}
\maketitle

\begin{abstract}
We provide a comprehensive investigation of different custom and off-the-shelf architectures as well as different approaches to generating feature vectors for offensive language detection. We also show that these approaches work well on small and noisy datasets such as on the Offensive Language Identification Dataset (OLID), so it should be possible to use them for other applications.
\end{abstract}

\section{Introduction}
During this exploration, we investigate different methods for tweet classification. In the first task, we label tweets as offensive or not. In the second we predict whether an offensive tweet is targeted at an individual or not. Finally, in the third task, we perform target identification by classifying targetted tweets as targeted at a group or at an individual. All three tasks are classification problems to which we explored two main solution classes. Firstly, we computed feature vectors - using naive methods, and then with the use of Word2Vec \cite{WORD2VEC}. We later fed those feature representations into simple scikit-learn methods. Secondly, we moved on to end-to-end machine learning methods that were able to classify without extracting vector features beforehand. We compared performance of CNN and LSTM as well as effectiveness of BERT \cite{bert} and FastText \cite{joulin2016bag} models. 

\section{Methods}

\subsection{Dataset}

The provided dataset (OLID) \cite{OLID, offenseval} is split into four parts, training data and testing data for task A, B, C. The training dataset contains a training and trial (validation) set with 13,240 and 320 tweets respectively. Each of these tweets has three different labels corresponding to the correct output for each of the tasks. 

\subsection{Preprocessing}

Given that the dataset has been partially preprocessed this task was simpler. In the training set, all usernames have been replaced by \texttt{@USER} and all URLs by \texttt{URL}. However, it still contained non-alphanumeric symbols such as emojis, all of these were removed only keeping letters, numbers and punctuation while converting all letters to lower-case. We also expanded all word contraction (eg. don't was expanded into two words: do not), this will make it easier for our models to recognize negation as all of them are word-based instead of character-based.

\subsection{Feature vectors and simple classification}

\subsubsection{Naive feature vectors}

The first features which we created using a naive approach of counting the number of occurrences of hate\footnote{https://hatebase.org/}, positive and negative \cite{positive-negative-1, positive-negative-2} words in a tweet. This produces three features, three more features were generated by normalizing the occurrence of hate, positive and negative words by the length of the tweet.

\subsubsection{Word2Vec}
\label{Word2Vec}
Word2Vec \cite{WORD2VEC} is a machine learning model for generating word embeddings. Word vectors are positioned in the vector space such that words that share common contexts in the corpus are located in close proximity to one another \cite{WORD2VEC}. Such vectors can be generated based on pre-trained models (scraped Wikipedia pages) or just using training text data. We fine-tuned our model by doing a random search on the number of dimensions and width of the context window, and found that using a pre-trained model resulted in inferior performance.  We then experimented with different means (averaging, taking a max, etc...) of transforming vectors for multiple words in a tweet to a single vector so that the feature representation could be tested with various scikit-learn classification methods (\ref{skcit}).

\subsubsection{Simple classifiers}
\label{skcit}
As a baseline we used RandomForestClassifier, GradientBoostingClassifier, LogisticRegression and MLPClassifier from the scikit-learn library \cite{scikit-learn}. All of these classifiers were trained separately with the default parameters. 

A VotingClassifier and a StackingCVClassifier were also compared. They combined all of the classifiers mentioned above to give a single result.

\subsection{CNN}
We implemented a Convolutional Neural Network based on the model presented in Kim's 
\textit{Convolutional Neural Networks for Sentence Classification} \cite{Yoon}. Each tweet is input as a 2D array of dimensions {$($\tt max\_tweet\_length $\times$ \tt embedding\_length$)$}. The embedded tweets are then passed to a convolutional layer containing three parallel convolutions with different filter sizes. The result of this layer is max-pooled and the three parallel results are concatenated into one long feature vector. A dropout layer is used for regularization, and then a fully connected softmax layer gives the resulting classification. This was implemented in Keras \cite{chollet2015keras} with a Tensorflow \cite{tensorflow2015-whitepaper} backend, and trained using the Adam optimiser and binary cross entropy loss. We used early stopping to keep the model that produced the best F1 score on the validation set during training. For the word embeddings, at first we tried using our word2vec embeddings for the tweets but with the CNN these did not work well, so instead, we added an embedding layer to the front of the network which is trained with the CNN.

\subsection{LSTM}

The LSTM model uses a very similar design to the CNN. It uses an embedding layer and three convolutional layers with a kernel size of 3, 4 or 5 multiplied by the embedding dimension. Compared to the CNN the MaxPooling layer is removed and the concatenated outputs of the convolutional layers are then fed into a bidirectional LSTM which returns a single output. Everything else is kept the same as in the CNN to allow for better comparison between the two different architectures.

We also tested another version of the LSTM which did not use the convolutional layers however this overfit very quickly and would always return the same class in most of the cases and it was not pursued any further.

\subsection{FastText}
We explored the suitability of FastText - a popular text classification library open-sourced by Facebook. It leverages n-grams features and other tricks described in \cite{joulin2016bag} to achieve high efficiency on large text corpus. We decided to train the model with the use of pre-trained vectors coming from scraped English Wikipedia pages and with no pre-computed features. We found that Wiki dictionary does not align with words used in tweets and as such experiments with pre-trained vectors consistently resulted in worse results for classification (similarly as in \ref{Word2Vec}). Thus, we continued on fine-tuning the model (without pre-training) with the use of random search. For each task we use random search to fine-tune: number of training epochs, n for wordNgrams, training learning rate, the minimum number of occurrences for a word to be included in the dictionary, width of the context window and word vector dimensionality. Each training loop takes only a couple of seconds on a CPU which allowed us to run a large number of experiments for each task. We present its results in \ref{Faexperiments}. In general, we found FastText to be very fast and perform well on small dataset classification, even though it has been designed for bigger datasets. 

\subsection{BERT}

BERT \cite{bert}, which stands for Bidirectional Encoder Representations from Transformers was developed by the Google AI language team. This model achieves state-of-the-art results on eleven natural language processing tasks. Auto-Keras \cite{bert} was used to train a pre-trained BERT representation which can be fine-tuned with just one additional output layer. Both base and large models were trained with and without case sensitivity. However we found it interesting that BERT performs well when it is initialized with weights however FastText does not. The best results were achieved with the base model without case sensitivity. We attribute this to the size of the fact that the dataset is relatively small with short sentences. Also given that many tweets are not syntactically correct and use capitals to indicate excitement instead of only capitalizing the first letter of a proper noun they introduce a lot of noise and it distinguishing the words base on the capitalization does not help the model. 

\section{Experiments}
\label{Faexperiments}

\subsection{Simple Classifiers}

First, we compared the two different feature vectors which we created, the naive one which counts the occurrence of hate, positive and negative words in a tweet and then the vectors generated by word2vec.

\begin{table}[h]
\begin{tabular}{l|l|l|l|}
\cline{2-4}
                                                                                             & Naive          & W2V            & Both           \\ \hline
\multicolumn{1}{|l|}{Logistic Regression}                                                    & \textbf{0.637} & 0.43           & 0.658          \\ \hline
\multicolumn{1}{|l|}{MLP Classifier}                                                         & 0.633          & 0.497          & \textbf{0.668} \\ \hline
\multicolumn{1}{|l|}{\begin{tabular}[c]{@{}l@{}}Random Forest\\ Classifier\end{tabular}}     & 0.565          & \textbf{0.537} & 0.589          \\ \hline
\multicolumn{1}{|l|}{\begin{tabular}[c]{@{}l@{}}Gradient Boosting\\ Classifier\end{tabular}} & 0.627          & 0.49           & 0.629          \\ \hline
\multicolumn{1}{|l|}{SVC}                                                                    & 0.630          & 0.432          & 0.632          \\ \hline
\multicolumn{1}{|l|}{VotingClassifier}                                                       & \textbf{0.637} & 0.496          & 0.665          \\ \hline
\multicolumn{1}{|l|}{\begin{tabular}[c]{@{}l@{}}Stacking CV \\ Classifier\end{tabular}}      & 0.626          & 0.536          & 0.657          \\ \hline
\end{tabular}
\caption{\label{tab:simple-classifiers} The macro F1 scores for Task A when using different simple classifiers evaluated on the trial set.}
\end{table}

The Voting Classifier and Stacking CV Classifier always use three classifiers which achieve the best results for the features used. In table \ref{tab:simple-classifiers} we can see that just using the occurrence of hate, positive and negative words gives us very good results, much better than using Word2Vec on its own however the fusion of both feature vectors gives the best results.

\begin{figure}
  \includegraphics[width=\linewidth]{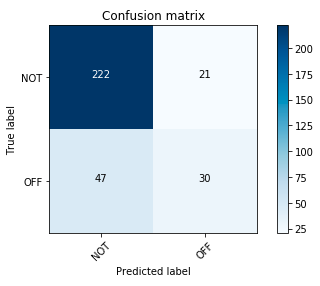}
  \caption{Confusion matrix for the MLP Classifier when using both feature vectors.}
  \label{fig:confusion-1}
\end{figure}

In figure \ref{fig:confusion-1} we can see the confusion matrix for the results with the highest macro F1 score using the MLP Classifier. It is visible that the high-class imbalance causes the false negative classification of offensive tweets to be very high however if we were to only look at the accuracy the results might make it seem that the model is very good.

\subsection{CNN}
When training the CNN we spent some time tuning hyperparameters to improve the architecture. We mainly varied parameters related to network architecture. This included the embedding dimension, convolution filter sizes and number of filters and the dropout parameter. Changing these parameters did not lead to a significant change in network performance. Finally, we settled on the hyperparameters shown in table \ref{tab:CNN-hyparams}. Performance with these parameters on the validation set and test set can be seen in table \ref{tab:validation-set} and table \ref{tab:test-set} respectively. The performance on Task A is quite good, and Task C is not so bad when compared to other methods. However, on Task B this model completely failed to learn anything other than always predicting the most common class.
\begin{table}[h]
\begin{tabular}{l|l|l|l|}
\hline
\multicolumn{1}{|l|}{Parameter} & Value      \\ \hline
\multicolumn{1}{|l|}{Embedding size} & 256\\ \hline
\multicolumn{1}{|l|}{Num. filters} & 768   \\ \hline 
\multicolumn{1}{|l|}{Filter sizes} & 3, 4, 5    \\ \hline 
\multicolumn{1}{|l|}{Dropout parameter} & 0.5 \\ \hline
\multicolumn{1}{|l|}{Batch size} & 30\\ \hline
\multicolumn{1}{|l|}{Adam Learning Rate} & $1 \times 10^{-4}$ \\ \hline
\multicolumn{1}{|l|}{Adam Beta 1} & 0.9 \\ \hline
\multicolumn{1}{|l|}{Adam Beta 2} & 0.999 \\ \hline
\multicolumn{1}{|l|}{Adam Epsilon} & $1 \times 10^{-8}$ \\ \hline
\end{tabular}
\caption{\label{tab:CNN-hyparams} Hyperparameters used on the Convolutional Neural Network model.}
\end{table}
\subsection{LSTM}

Even though the architecture of the LSTM which we trained was very similar to the CNN it proved to be very challenging to train. The model was over-fitting to the training data and even after extensive tweaking of the hyperparameters such as learning rate, number of filters, dropout and recurrent dropout of the LSTM layer we were not able to achieve better results than in the CNN. Due to the time constraints, we switched our focus to the other models mentioned.

\subsection{BERT and FastText}

Both BERT and FastText were relatively easy to use and didn't require as much tweaking as the other approaches used. In this case, most of the time was spent on devising a good method for preprocessing the tweets and then integrating the models without training pipeline.

\subsection{Comparison of models}

For the final submission, we used the CNN which we built as well as FastText and BERT because these achieved the best results during training. The macro F1 scores for the validation set can be seen in table \ref{tab:validation-set} and the scores for the test set in table \ref{tab:test-set}.

\begin{table}[h]
\begin{tabular}{l|l|l|l|}
\cline{2-4}
                               & Task A & Task B & Task C \\ \hline
\multicolumn{1}{|l|}{CNN}      & 0.786  & 0.330  & 0.381  \\ \hline
\multicolumn{1}{|l|}{LSTM}     & 0.753  & 0.330  & 0.381  \\ \hline
\multicolumn{1}{|l|}{FastText} & 0.785  & 0.645  &  \textbf{0.428} \\ \hline
\multicolumn{1}{|l|}{BERT}     & \textbf{0.824}  & \textbf{0.680}  & 0.375  \\ \hline
\end{tabular}
\caption{\label{tab:validation-set} The macro F1 scores on the validation set.}
\end{table}

\begin{table}[h]
\begin{tabular}{l|l|l|l|}
\cline{2-4}
                               & Task A & Task B & Task C \\ \hline
\multicolumn{1}{|l|}{CNN}      & 0.735 & 0.470  & 0.457 \\ \hline
\multicolumn{1}{|l|}{FastText} & 0.707 & 0.560 & 0.319 \\ \hline
\multicolumn{1}{|l|}{BERT}     & \textbf{0.786} & \textbf{0.687} & \textbf{0.561} \\ \hline
\end{tabular}
\caption{\label{tab:test-set} The macro F1 scores on the test set.}
\end{table}

From this, we can see that BERT generalized better to the test dataset compared to our other models. We find it very interesting that the results for FastText were worse than the CNN for Task A and Task C as it performed better for the validation set. We assume that this is because we performed extensive hyperparameter search on the training and validation set which might have caused it to overfit.

\section{Conclusion}

We have explored various architectures for text classification. Calculating vector features representations and feeding into scikit-learn classifiers proved to under-perform against end-to-end machine learning methods. Word2Vec, in general, also underperformed compared to naive features (probably due to the size of the training set and tweet-specific language in it), but a combination of two yielded a better result. 

However, in the end, we found that building a convolutional model or using a state-of-art model which we just modified for this task yielded the best results.
\onecolumn{
\bibliographystyle{splncs04_trunc}
\bibliography{cites}
}

\end{document}